\documentclass[runningheads]{llncs}
\usepackage[T1]{fontenc}
\usepackage{graphicx}
\usepackage{algorithm}
\usepackage{amsmath}
\usepackage{algpseudocode}
\usepackage{placeins}
\usepackage{xcolor}
\usepackage{hyperref}
\usepackage{subcaption}
\begin{document}

\title{DGG-XNet: A Hybrid Deep Learning Framework for Multi-Class Brain Disease Classification with Explainable AI}
%
%
\author{
Sumshun Nahar Eity\inst{1}\orcidID{0009-0003-4614-3883} 
\and Mahin Montasir Afif\inst{1}\orcidID{0009-0003-8499-2624}
\and Tanisha Fairooz\inst{1}\orcidID{0009-0004-0898-1381}
\and Md. Mortuza Ahmmed\inst{1}\orcidID{0000-0002-4735-571X}
\and Md Saef Ullah Miah\inst{1}\orcidID{0000-0003-4587-4636}}
\authorrunning{S. N. Eity et al.}

\institute{
American International University-Bangladesh, Dhaka-1229, Bangladesh
\email{22-48227-2@student.aiub.edu, 22-46573-1@student.aiub.edu, 22-49347-3@student.aiub.edu, mortuza@aiub.edu, saef@aiub.edu}\\
}
\maketitle              
\begin{abstract}
Accurate diagnosis of brain disorders such as Alzheimer's disease and brain tumors remains a critical challenge in medical imaging. Conventional methods based on manual MRI analysis are often inefficient and error-prone. To address this, we propose DGG-XNet, a hybrid deep learning model integrating VGG16 and DenseNet121 to enhance feature extraction and classification. DenseNet121 promotes feature reuse and efficient gradient flow through dense connectivity, while VGG16 contributes strong hierarchical spatial representations. Their fusion enables robust multiclass classification of neurological conditions. Grad-CAM is applied to visualize salient regions, enhancing model transparency. Trained on a combined dataset from BraTS 2021 and Kaggle, DGG-XNet achieved a test accuracy of 91.33\%, with precision, recall, and F1-score all exceeding 91\%. These results highlight DGG-XNet's potential as an effective and interpretable tool for computer-aided diagnosis (CAD) of neurodegenerative and oncological brain disorders.

\keywords{Alzheimer’s disease \and Brain Tumors \and DenseNet121 \and VGG16 \and DGG-XNet \and Grad-CAM\and }
\end{abstract}

\section{Introduction}
Brain disorders such as Alzheimer's disease together with brain tumors are substantial global health issues. Millions of patients are impacted, and medical practitioners face challenges in diagnosing them \cite{10.1007/978-981-96-3294-7_26}. Studies have demonstrated that, depending on the patient's age and the brain region involved, 52\% of patients had Alzheimer's Disease Neuropathologic Change (ADNC) in the glioblastoma-adjacent cortex tissue \cite{greutter2024frequent}. According to research, there is presently a demand for enhanced computational diagnostics to increase diagnostic speed due to the rising number of people with Alzheimer's disease (246 million globally, 39 million of whom are critical) and brain tumors. According to the World Health Organization \cite{who2024neurological}, since over 3 billion individuals worldwide suffer from various brain diseases, early detection methods become crucial.

In order to detect diseases medical personnel perform interpretive analysis of Magnetic Resonance Imaging (MRI) which scans the human body, even though it's a revolutionary invention it is still a slow process and contains manual errors. Although in the current circumstances, MRI is the only leading medical tool for discovering brain diseases that involve neurodegeneration and cancer disorders. However, acquiring the MRI report requires people to see the doctor again to obtain the findings, which is both costly and time-consuming. For better medical outcomes, timely, accurate diagnosis requires both accuracy and efficiency, as evidenced by the number of diagnosed patients.

The goal of this research is the development of an efficient and quick detection model using MRI technology in recognition of Alzheimer’s disease and brain tumors at an early stage. In combining advanced imaging methods with machine learning, this study hopes to improve diagnostic accuracy and thus patient outcomes and treatment plans.

CNNs have shown remarkable effectiveness in picture classification tasks, single designs usually have trouble applying effectively across a range of neurological disorders. To face this circumstance, this investigation presents DGG-XNet, it is a novel hybrid deep learning model based on VGG16 and DenseNet121, which fully utilize their characteristic extractions. Regarding DenseNet121 \cite{8099726}, its feature reuse is greatly improved by dense connections generated via the usage of dense layers, while VGG16's \cite{simonyan2015very} deep hierarchical design keeps robust spatial characteristics useful for medical image-related jobs fiercely. DGG-XNet is designed to improve the multi-class classification efficiency of brain tumors and Alzheimer's disease by combining these two models.

\section{Literature Review}
\label{lr}
\subsection{Alzheimer's disease}
Pursuit of Alzheimer's disease (AD) diagnosis becomes an increasing fear for persons as they age.  The cellular breakdown from Alzheimer's disease leads people to lose memory of familiar people along with the ability to learn new data or recognize their relatives \cite{jahn2013memory}. The disease affects the ability to determine which direction people should go and their ability to recognize the environment.  Eating is combined with breathing and coughing as abilities that fade away during the advanced stages of disease progression. AD and VD, share sufficient overlapping symptoms that make diagnosing AD particularly challenging both in behavioral and psychiatric domains as well as in language and memory dysfunction \cite{huang2020altered,castellazzi2020machine}. Early and accurate identification of AD and monitoring of its progression will greatly benefit patient care as well as treatment and prevention.

\subsection{Brain Tumor}
A brain tumor stands as an extremely dangerous medical condition that threatens the safety of the brain. Brain tumors develop regularly because the existing vein and nerve damage in the brain leads to this condition.  The progression of tumor growth determines whether the patient loses partial or complete sight to this condition \cite{zaw2019brain}. The development of brain tumors also depends on serious myopia cases as well as ethnic characteristics and genetic background \cite{ghnemat2023ischemic}. The condition is caused by insufficient blood circulation and prevents new nerve blood vessels from growing.  Rapid and automated, the discovery of early diagnostic techniques has become crucial in modern developed societies because of their advanced requirements.

\subsection{Deep Learning and Brain Diseases}
Deep learning techniques have become widely used in medical image classification by providing effective approaches for classifying diseases in brain MRI scans. A study by \cite{ghazal2022alzheimer} shows an automated system for the early detection of Alzheimer's disease using transfer learning on multi-class classification of brain MRI images. The proposed model achieved 91.70\% accuracy. Similar study by \cite{SHARMA2022100506} using two MRI datasets-containing 6400 and 6330 images-employed a neural network classifier with VGG16 as the feature extractor. The model achieved accuracy, precision, recall, AUC, and F1-score of 90.4\%, 0.905, 0.904, 0.969, and 0.904, respectively, on the first dataset, and 71.1\%, 0.71, 0.711, 0.85, and 0.71 on the second dataset. This study by \cite{poma2020optimization} shows an approach to optimize CNNs using the Fuzzy Gravitational Search Algorithm, focusing on key parameters like image block size, filter number, and filter size. Applied to the ORL and Cropped Yale databases, the optimized CNN outperformed non-optimized models. In AD classification, VoxCNN achieved 79\% accuracy and 88\% AUC, while ResNet achieved 80\% accuracy and 87\% AUC. 

A research by \cite{RAMMURTHY20223259} proposes a WHHO-based DeepCNN model for brain tumor detection using MRI images. It combines Whale Optimization and Harris Hawks Optimization with DeepCNN for segmentation and feature extraction. The model achieved 0.816 accuracy, 0.791 specificity, and 0.974 sensitivity, outperforming other methods. Another study by \cite{lerousseau2020multimodal} investigates a deep learning-based approach for tumor classification, combining whole slide images and MRI scans. The model was evaluated on the 2020 Computational Precision Medicine Challenge in a 3-class unbalanced classification task. It achieved impressive performance with a balanced accuracy of 0.913. Another research paper by \cite{8748288} proposes an automatic brain tumor detection method using genetic algorithm for image s and support vector machine (SVM) for classification of brain MRI images. The model uses CNN with Discrete Wavelet Transform (DWT) and classifies tumors as Normal vs. Not Normal, achieving 85\% accuracy.

Although a good number of work has been done in multi-class brain disease classification, challenges such as accuracy, dataset imbalance, and identifying the most reliable regions of images for classification tasks still remain. To address these issues, this study proposes the DGG-XNet model, which combines VGG16 and DenseNet121 to enhance accuracy. Additionally, the model incorporates two Explainable AI (XAI) techniques including Grad-CAM and Integrated Gradients, to make its predictions more transparent and interpretable, thereby improving its practical applicability in real-world medical imaging. Table \ref{tab:paper_comparison}  presents a comparison of accuracy among discussed existing works.

\begin{table}[htbp]
\centering
\caption{Accuracy Comparison of Previous Works}
\label{tab:paper_comparison}
\begin{tabular}{@{}|l| l| l| c|@{}}  
\hline
Reference & Model & Dataset & Accuracy \\
\hline
\cite{ghazal2022alzheimer} & Transfer learning & MRI images & 91.70\% \\ \hline
\cite{SHARMA2022100506} & VGG16+feature extractor & 2 MRI datasets & 90.4\%, 71.1\% \\ \hline
\cite{poma2020optimization} & VoxCNN, ResNet & ORL, Cropped Yale & 79\%, 80\% \\ \hline
\cite{RAMMURTHY20223259} & WHHO+DeepCNN & MRI images & 81.6\% \\ \hline
\cite{lerousseau2020multimodal} & Ensemble model & TCIA+CBICA MRI images & 91.3\% \\ \hline
\cite{8748288} & CNN+DWT, SVM & Brain Tumor Dataset & 85\% \\
\hline
\end{tabular}
\end{table}

\section{Methodology}

\begin{figure}[htbp]
    \centering
    \includegraphics[width=.5\linewidth]{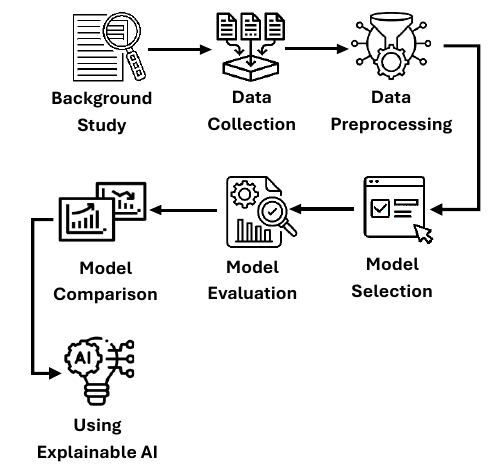} 
    \caption{Methodology Overview }
    \label{fig:methodology}
\end{figure}

In this section, we describe our proposed methodology. An extensive review of earlier studies in the area of neuroimaging-based disease classification was conducted before this study. Based on the insights, a dual-path hybrid model was constructed and trained on a balanced dataset. The methodology pipeline is summarized in Fig. \ref{fig:methodology}.

\subsection{Dataset Description}
The study utilized two publicly available datasets. By combining “BraTS 2021 Task 1 dataset” \cite{baid2021rsna,menze2015multimodal,bakas2017advancing} which contains MRI volumes for brain tumor classification and “alzheimers-dataset-4-class-of-images” which contains 2D axial brain images across four categories: Non-Demented, Very Mild Demented, Mild Demented, and Moderate Demented from Kaggle were collected and merged. 

\subsection{Data Preprocessing}
Images were resized to 224×224×3 and normalized for model compatibility. From BraTS 2021 Task 1, Only the T1-weighted modality was used in this study. 2D slices were extracted from the 3D volumes by selecting evenly spaced axial slices to preserve critical spatial information. For the Alzheimer’s dataset, images were pre-processed and resized uniformly. To prevent model bias due to class imbalance, the dataset was equalized using down sampling. The number of samples in each class was adjusted to match the minority class, resulting in 500 samples per class. 

\subsection{Dataset Split}
The dataset was partitioned into training (70\%), validation (20\%), and testing (10\%) sets using stratified sampling. This ensured that each subset retained balanced class distribution. Table \ref{tab:dataset_split} presents the distribution of the dataset in different sets.

\begin{table}[htbp]
\centering
\caption{Dataset Distribution }
\begin{tabular}{|l|c|c|c|}
\hline
Class& Train& Validation& Test\\
\hline
Tumour         & 350            & 100                 & 50            \\ \hline
Normal         & 350            & 100                 & 50            \\ \hline
Alzheimer’s    & 350            & 100                 & 50            \\
\hline
Total& 1050           & 300                 & 150           \\
\hline
\end{tabular}
\label{tab:dataset_split}
\end{table}

\subsection{Proposed Model}
This study proposes a hybrid deep learning architecture that combines the feature extraction power of two widely used convolutional neural networks: VGG16 and DenseNet121. Both models were initialized with pre-trained ImageNet \cite{5206848} weights to benefit from transfer learning, and their convolutional layers were initially frozen to retain learned representations.

Each input image is passed through both VGG16 and DenseNet121 backbones. Feature maps are extracted using Global Average Pooling \cite{lin2013network} from each model, and then concatenated to form a unified feature vector:

\begin{equation}
F = \text{Concat}\left(\text{GAP}(\text{VGG16}(x)),\ \text{GAP}(\text{DenseNet121}(x))\right)
\label{eq:concat_features}
\end{equation}
where $x$ is the input image and $F$ is the fused feature vector.

The merged features are passed through a series of fully connected layers with ReLU activation \cite{agarap2018deep} functions and dropout for regularization. The final output layer uses the softmax activation \cite{10057747} to classify the input into one of three categories: Tumour, Normal, or Alzheimer’s.

\begin{equation}
\hat{y} = \text{softmax}(W_2 \cdot \text{ReLU}(W_1 \cdot F + b_1) + b_2)
\label{eq:classification_head}
\end{equation}

The model is compiled using the Adam optimizer \cite{kingma2014adam} with a learning rate of 0.0001. Adam is an adaptive optimizer that updates parameters using first and second moment estimates of the gradients:

\begin{equation}
\theta_t = \theta_{t-1} - \frac{\alpha}{\sqrt{\hat{v}_t} + \epsilon} \cdot \hat{m}_t
\label{eq:adam_update}
\end{equation}
where $\theta_t$ is the model parameter at time step $t$, $\hat{m}_t$ and $\hat{v}_t$ are the bias-corrected estimates of the first and second moments, and $\epsilon$ is a small constant for numerical stability.

The categorical crossentropy \cite{zhang2018generalized} loss function was used to optimize the model, defined as:

\begin{equation}
\mathcal{L}_{\text{CCE}} = -\sum_{i=1}^{C} y_i \log(\hat{y}_i)
\label{eq:crossentropy}
\end{equation}
where $C$ is the number of classes, $y_i$ is the true label, and $\hat{y}_i$ is the predicted probability for class $i$.

\begin{figure}[htbp]
    \centering
    \includegraphics[width=0.95\linewidth]{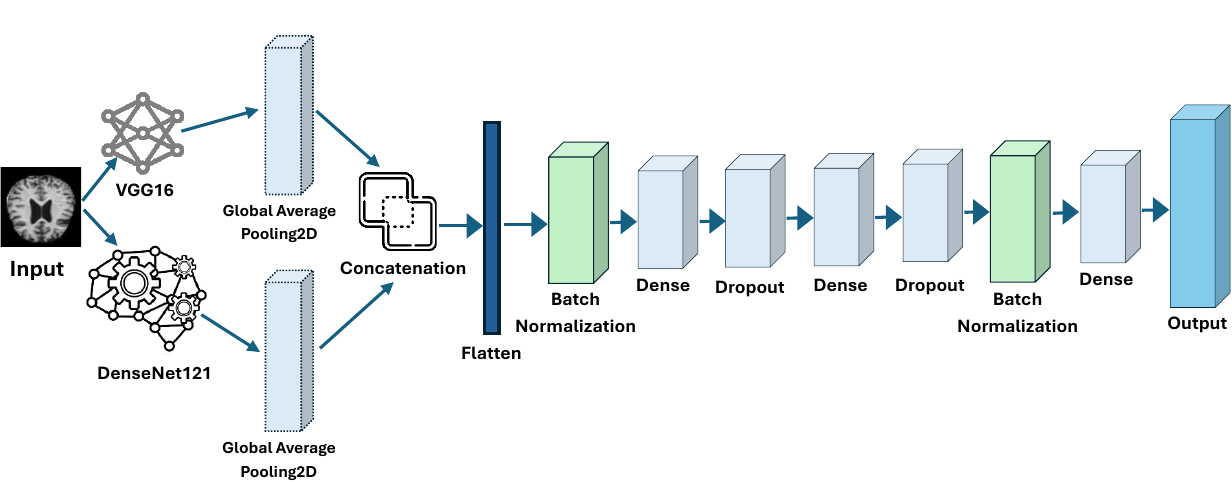}
    \caption{Architecture of the proposed model}
    \label{fig:hybrid_architecture}
\end{figure}

To prevent overfitting and reduce unnecessary computation, EarlyStopping technique was applied during training. This monitors the validation loss and restores the best weights when no improvement is observed after 5 consecutive epochs. Figure \ref{fig:hybrid_architecture} presents the architecture of the proposed model.

\subsection{Model Evaluation Metrics}
To evaluate the model’s performance, several commonly used classification metrics were applied. These include accuracy, precision, recall, and F1-score. Each metric provides insight into different aspects of model performance.

\subsubsection{Accuracy}
Accuracy measures the proportion of total correct predictions made by the model:

\begin{equation}
\text{Accuracy} = \frac{TP + TN}{TP + TN + FP + FN}
\end{equation}

\subsubsection{Precision}

Precision shows how many of the predicted positive samples were actually correct:

\begin{equation}
\text{Precision} = \frac{TP}{TP + FP}
\end{equation}

\subsubsection{Recall}
Recall (or sensitivity) measures how many actual positive samples were correctly identified:

\begin{equation}
\text{Recall} = \frac{TP}{TP + FN}
\end{equation}

\subsubsection{F1-score}
F1-score is the harmonic mean of precision and recall. It provides a balanced metric, especially useful when classes are imbalanced:

\begin{equation}
\text{F1-score} = 2 \times \frac{\text{Precision} \times \text{Recall}}{\text{Precision} + \text{Recall}}
\end{equation}

Here, $TP$, $TN$, $FP$, and $FN$ represent true positives, true negatives, false positives, and false negatives, respectively.

\subsection{Explainable AI (XAI)}
To better understand and interpret the decisions made by the proposed hybrid model, Explainable AI (XAI) \cite{5206848} techniques were utilized. These techniques help highlight the regions of the input image that influenced the model’s predictions. In this study, Grad-CAM and Integrated Gradients were used to generate visual explanations.

\subsubsection{Grad-CAM}
Gradient-weighted Class Activation Mapping (Grad-CAM) \cite{8237336} is a widely used technique to visualize the important regions in an image that contribute to a model’s decision. Grad-CAM works by computing the gradient of the predicted class score with respect to the feature maps of the final convolutional layer. These gradients are then used to create a coarse heatmap that highlights discriminative regions.

The Grad-CAM heatmap $L_{\text{Grad-CAM}}^c$ for class $c$ is computed as:

\begin{equation}
L_{\text{Grad-CAM}}^c = \text{ReLU} \left( \sum_k \alpha_k^c A^k \right)
\label{eq:gradcam}
\end{equation}

where $A^k$ represents the $k$-th feature map, and $\alpha_k^c$ is the weight computed by global average pooling over gradients:

\begin{equation}
\alpha_k^c = \frac{1}{Z} \sum_i \sum_j \frac{\partial y^c}{\partial A_{ij}^k}
\end{equation}

Here, $y^c$ is the score for class $c$, and $Z$ is the number of pixels in the feature map. The ReLU function ensures only positive influences are visualized.

\subsubsection{Integrated Gradients}

Integrated Gradients\cite{zhuo2024ig2} was also employed to gain finer, pixel-level insights into the model's decisions. This method attributes the prediction of a model to its input features by integrating the gradients of the output with respect to the input along a straight path from a baseline (usually a black image) to the actual input.

Formally, the integrated gradient for the $i$-th input feature is given by:

\begin{equation}
\text{IG}_i(x) = (x_i - x_i') \times \int_{\alpha=0}^1 \frac{\partial F(x' + \alpha (x - x'))}{\partial x_i} d\alpha
\end{equation}

where $x$ is the input, $x'$ is the baseline, and $F$ is the model's output function. This approach helps understand which pixels in the image most significantly contributed to the prediction.

\noindent The combination of Grad-CAM and Integrated Gradients provides both coarse and fine-grained visual explanations, enhancing model transparency and building trust in clinical applications.

\section{Results and Discussion}
The performance of the proposed DGG-XNet model was evaluated on a multi-class brain MRI classification task involving three classes: Tumour, Normal, and Alzheimer’s. The model achieved a test accuracy of  91.33\%, significantly outperforming several well-established deep learning architectures.

Table \ref{tab:model_comparison} presents a comparison between DGG-XNet and other popular CNN-based models, including VGG16, DenseNet121, MobileNetV2, InceptionV3, ResNet variants, and EfficientNetB3. Among these, VGG16 was the closest in performance, achieving an accuracy of 84.67\%. Other models showed moderate to lower performance, with ResNet101 recording the lowest at 68.00\%.

The strong performance of DGG-XNet can be attributed to the hybrid fusion of VGG16 and DenseNet121 feature extractors, enabling the model to capture both local and global patterns effectively. Additionally, techniques such as transfer learning, fine-tuning, data balancing, and the use of early stopping helped improve generalization while avoiding overfitting.

\begin{table}[htbp]
\centering
\caption{Model Accuracy Comparison}
\label{tab:model_comparison}
\begin{tabular}{|l|c|}
\hline
\textbf{Model}& \textbf{Accuracy}\\
\hline
\textbf{DGG-XNet (proposed model)}& \textbf{91.33\%}\\ \hline
VGG16           & 84.67\% \\ \hline
DenseNet121     & 82.67\% \\ \hline
InceptionV3     & 82.00\% \\ \hline
MobileNetV2     & 81.33\% \\ \hline
ResNet50        & 78.00\% \\ \hline
EfficientNetB3  & 74.67\% \\ \hline
ResNet101       & 68.00\% \\
\hline
\end{tabular}
\end{table}

The results confirm that DGG-XNet offers a more accurate and reliable solution for brain MRI classification, and demonstrates the effectiveness of hybrid feature fusion in complex medical imaging tasks.
\subsection{Training and Validation Performance}
\begin{figure}[htbp]
    \centering
    \includegraphics[width=\linewidth]{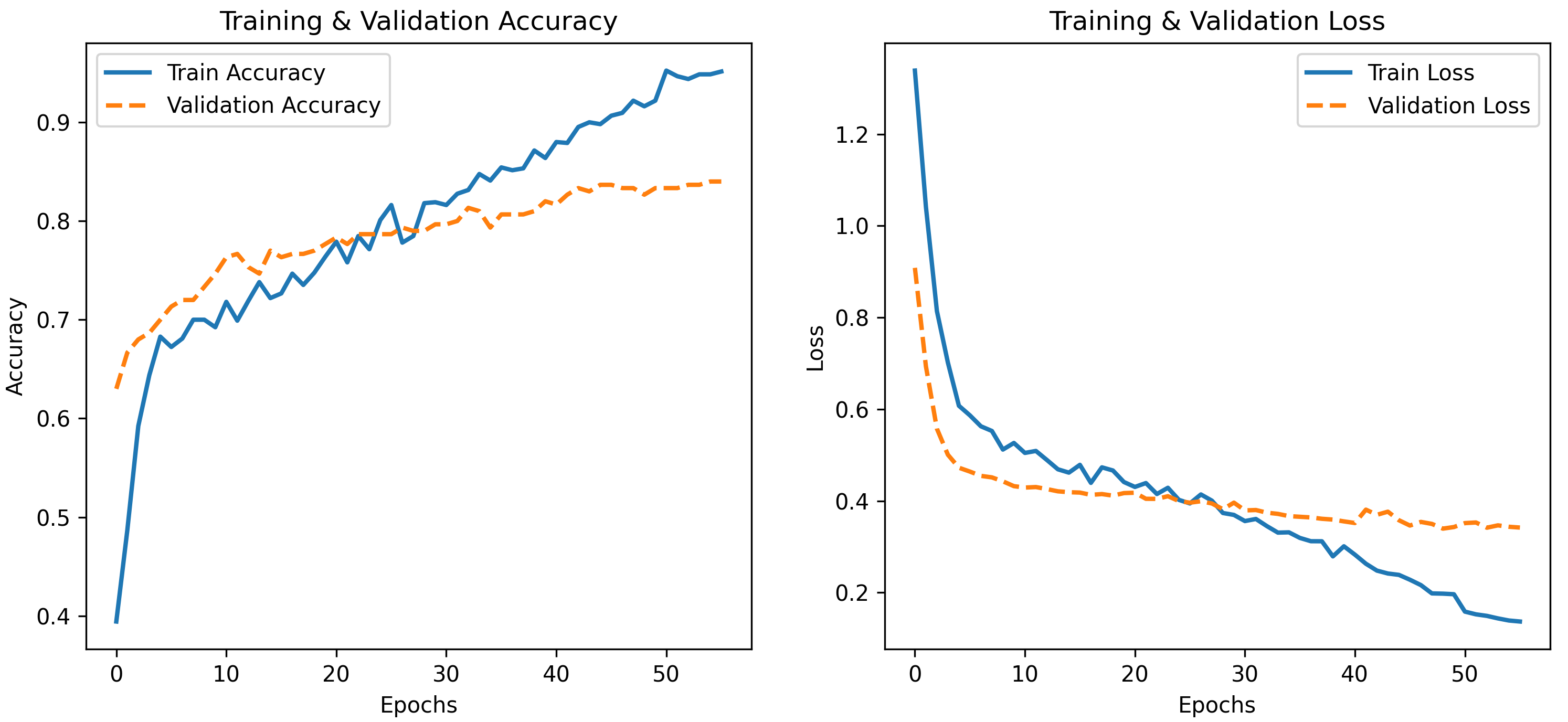}  
    \caption{Training and validation accuracy and loss curves.}
    \label{fig:train_val}
\end{figure}

Figure~\ref{fig:train_val} shows the training and validation accuracy and loss curves for the proposed model. The curves indicate that the model was able to learn effectively, with both training and validation accuracy gradually increasing and validation loss stabilizing, suggesting minimal overfitting.

\subsection{ROC Curve Analysis}
\begin{figure}[htbp]
    \centering
    \begin{subfigure}[b]{0.5\linewidth}
        \centering
        \includegraphics[width=\linewidth]{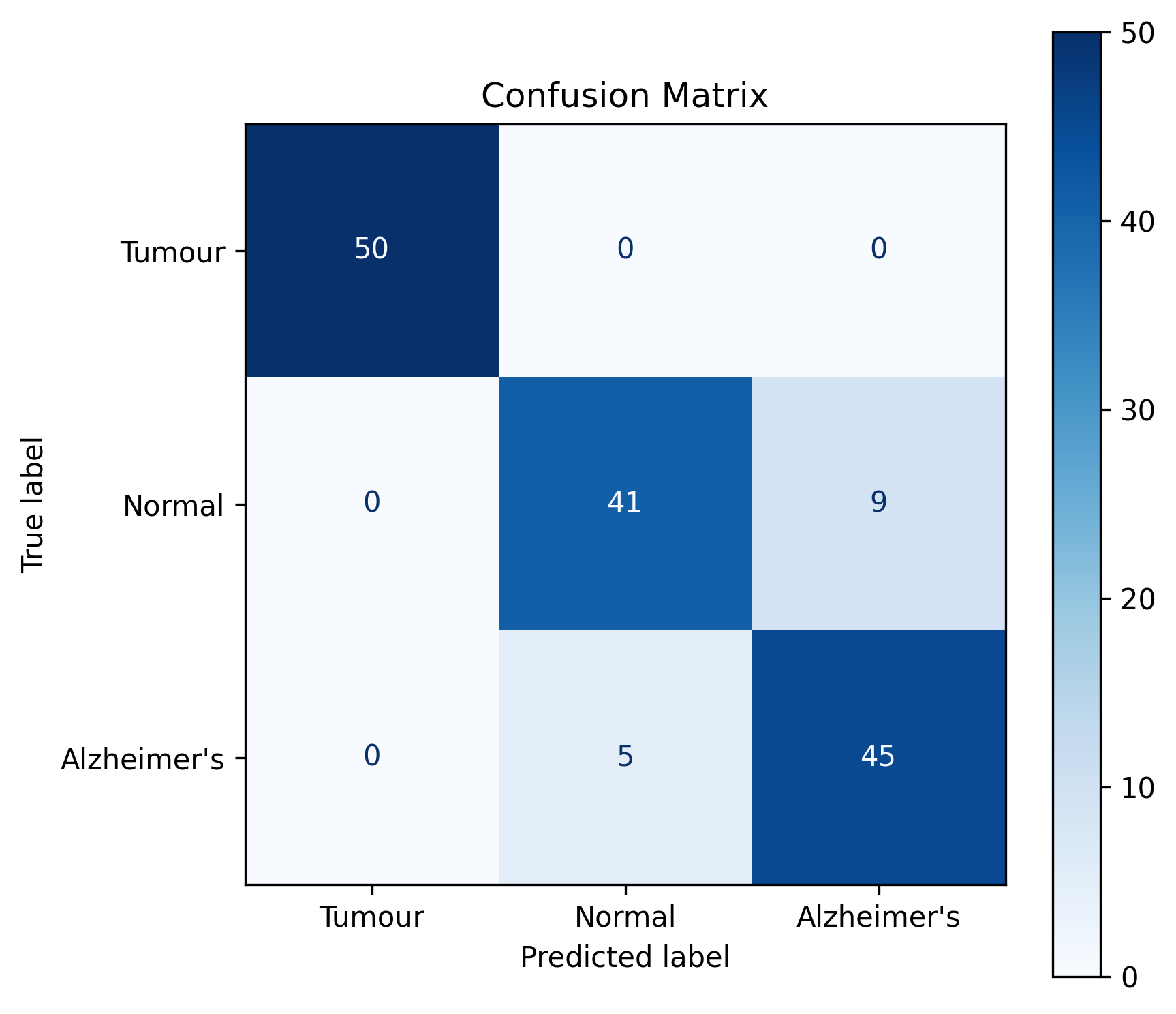}
        \caption{Confusion Matrix}
        \label{fig:conf}
    \end{subfigure}
    \hfill
    \begin{subfigure}[b]{0.48\linewidth}
        \centering
        \includegraphics[width=\linewidth]{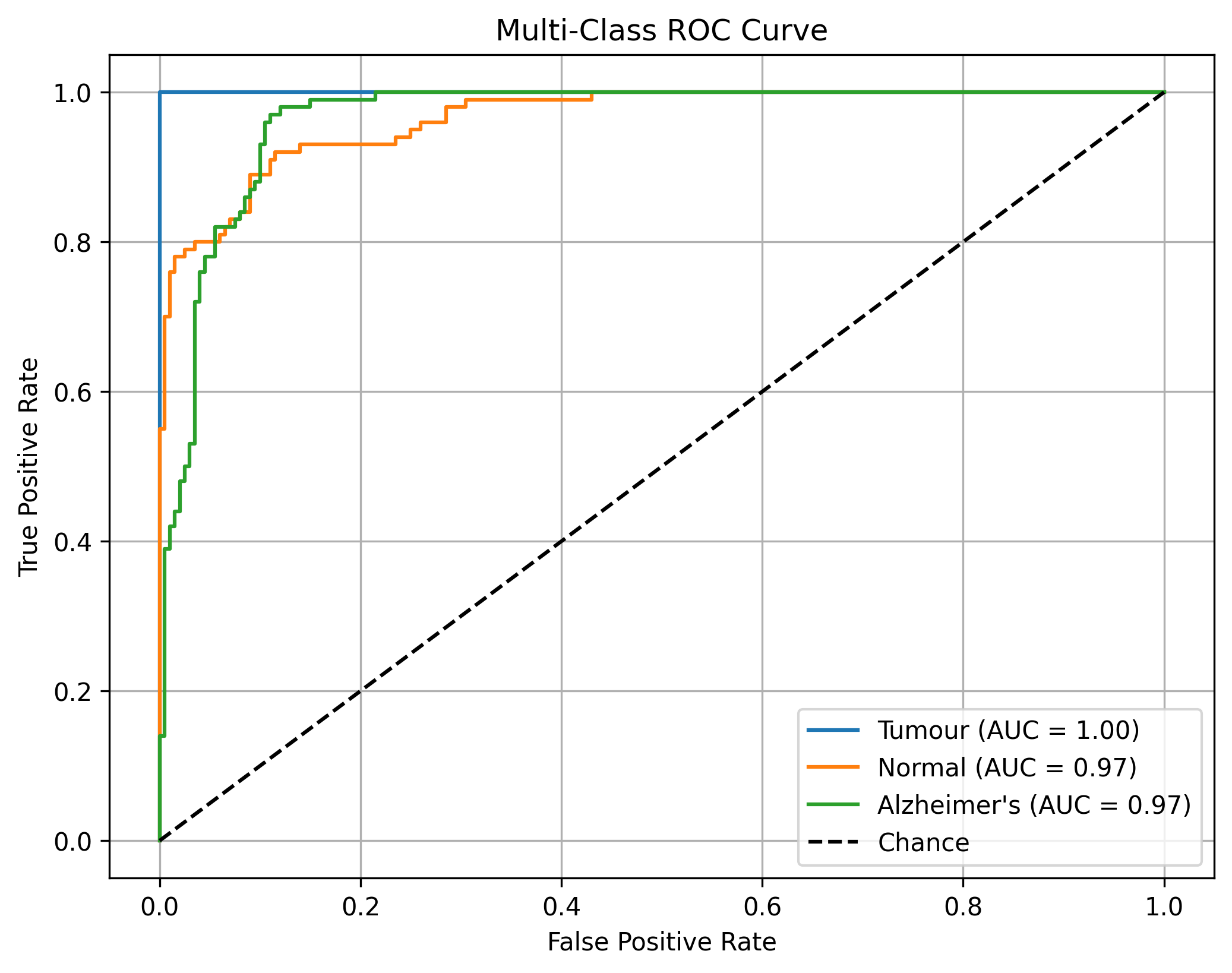}
        \caption{Multi-class ROC Curve}
        \label{fig:roc}
    \end{subfigure}
    \caption{Performance evaluation: Confusion Matrix and ROC Curve}
    \label{fig:combined}
\end{figure}

To evaluate the model’s ability to distinguish between the classes, ROC curves were plotted for each class using one-vs-rest classification. As shown in Figure \ref{fig:roc}, the proposed model achieved high AUC scores for all three classes, confirming its strong classification performance.

\FloatBarrier

\subsection{Confusion Matrix}
Figure~\ref{fig:conf} illustrates the confusion matrix based on a balanced test set (50 samples per class). The model demonstrated a high number of correct predictions across all classes, with very few misclassifications, further validating its effectiveness.

\subsection{Explainable AI Visualizations}
\begin{figure}[htbp]
    \centering
    \includegraphics[width=0.5\linewidth]{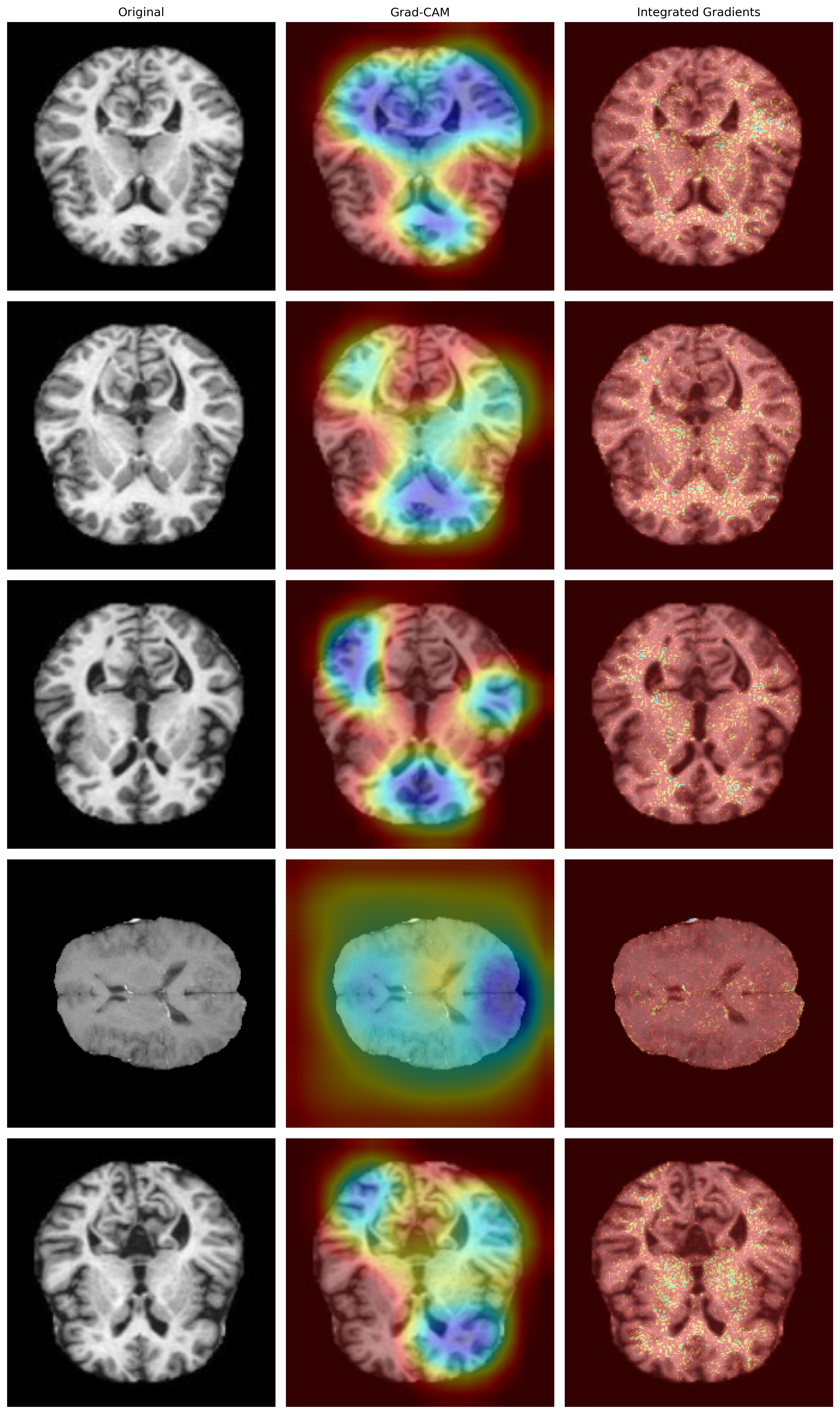}
    \caption{Explainable AI: Grad-CAM \& Integrated Gradients}
    \label{fig:xai}
\end{figure}

To gain insight into the decision-making process of the proposed model, explainable AI techniques such as Grad-CAM and Integrated Gradients were applied. Figure~\ref{fig:xai} displays these visualizations. Grad-CAM highlights the important regions of the image that influenced the model’s prediction, while Integrated Gradients provides pixel-level attribution. Together, these methods enhance the interpretability and transparency of the model, making it more trustworthy in clinical scenarios.

\subsection{Evaluation Metrics}

In addition to accuracy, precision, recall, and F1-score were computed to provide a more comprehensive evaluation of the proposed model. The results are summarized in Table~\ref{tab:metrics}, based on a test set of 150 samples. The model achieved an overall accuracy of 91.33\%, with a macro-average F1-score of  90.00\%.

\begin{table}[htbp]
\centering
\caption{Precision, Recall, and F1-score of the proposed model on the test set.}
\label{tab:metrics}
\begin{tabular}{|l|c|c|c|}
\hline
Class& Precision& Recall& F1-Score\\
\hline
Tumour       & 1.00 & 1.00 & 1.00 \\
Normal       & 0.89 & 0.87 & 0.88 \\
Alzheimer’s  & 0.82 & 0.84 & 0.83 \\
\hline
Macro Avg& 0.90 & 0.90 & 0.90 \\ \hline
Accuracy& \multicolumn{3}{|c|}{91.33\%} \\
\hline
\end{tabular}
\end{table}
\section{Conclusion}
This study proposed a hybrid deep learning model, DGG-XNet which was designed for multi-class classification of brain MRI images into Tumour, Normal, and Alzheimer’s categories. The architecture effectively combines the feature extraction strengths of VGG16 and DenseNet121, supported by global average pooling, feature fusion, and dense classification layers. Techniques such as transfer learning, class balancing, and fine-tuning were incorporated to improve generalization and robustness. The model achieved a high test accuracy of 91.33\%, outperforming several established deep learning architectures. Quantitative evaluation using precision, recall, and F1-score confirmed consistent performance across all classes. Furthermore, explainable AI techniques such as Grad-CAM and Integrated Gradients provided valuable visual insights into the model’s decision-making process, enhancing its transparency and trustworthiness in clinical contexts.

Future work may explore integrating 3D volumetric analysis, additional imaging modalities,  Data augmentation techniques, and domain adaptation to improve performance across varied datasets and real-world scenarios.
\bibliographystyle{splncs04}
\bibliography{references}

\end{document}